%% file: acl_camera_ready.tex
\pdfoutput=1

\documentclass[11pt]{article}

\usepackage{acl}

\usepackage{times}
\usepackage{latexsym}

\usepackage[T1]{fontenc}

\usepackage[utf8]{inputenc}

\usepackage{microtype}

\usepackage{multirow}
\usepackage{graphicx}
\usepackage{amsmath}
\usepackage{float}
\usepackage{bold-extra}
\usepackage{booktabs}
\usepackage{color}
\usepackage{hyperref}

\title{Attention Temperature Matters in Abstractive Summarization Distillation}

\author{
  Shengqiang Zhang$^1$ 
    \thanks{$^*$ Equal contribution.} 
    ~\thanks{$^\dag$ Work done during the authors' internships at Microsoft Research Asia.}, Xingxing Zhang$^2$$^*$, Hangbo Bao$^2$$^\dag$, Furu Wei$^2$
\\
  $^1$ Peking University \\
  $^2$ Microsoft Research Asia \\
  \texttt{sq.zhang@pku.edu.cn} \\
  \texttt{\quad \{xizhang,t-habao,fuwei\}@microsoft.com} \\
}

\begin{document}
\maketitle
\begin{abstract}
    Recent progress of abstractive text summarization largely relies on large pre-trained sequence-to-sequence Transformer models, which are computationally expensive. This paper aims to distill these large models into smaller ones for faster inference and with minimal performance loss. Pseudo-labeling based methods are popular in sequence-to-sequence model distillation. In this paper, we find simply manipulating attention temperatures in Transformers can make pseudo labels easier to learn for student models. Our experiments on three summarization datasets show our proposed method consistently improves vanilla pseudo-labeling based methods. Further empirical analysis shows that both pseudo labels and summaries produced by our students are shorter and more abstractive.
    Our code is available at~\url{https://github.com/Shengqiang-Zhang/plate}.
\end{abstract}

\section{Introduction} \label{INTRO}
Automatic document summarization is the task of rewriting a long document into its shorter form while still retaining its most important content. In the literature, there are mainly two kinds of methods for summarization: \emph{extractive summarization} and \emph{abstractive summarization} \cite{Nenkova:McKeown:2011}. In this work, we focus on abstractive summarization, which is viewed as a sequence-to-sequence (Seq2Seq) learning problem, since recent abstractive models outperform their extractive counterparts and can produce more concise summaries \cite{raffel2020exploring,lewis-etal-2020-bart,zhang2020pegasus,liu-lapata-2019-text}. Recent progress of abstractive summarization largely relies on large pre-trained Transformer models \cite{raffel2020exploring,lewis-etal-2020-bart,zhang2020pegasus,liu-lapata-2019-text,bao2020unilmv2}. With these extremely large models, we can obtain state-of-the-art summarization results, but they are slow for online inference, which makes them difficult to be used in the production environment even with cutting-edge hardware. 
This paper aims to distill these large Transformer summarization models into smaller ones with minimal loss in performance.


Knowledge distillation is a class of methods that leverage the output of a (large) teacher model to guide the training of a (small) student model. In classification tasks, it is typically done by minimizing the distance between the teacher and student predictions \cite{hinton2015distilling}. As to Seq2Seq models, an effective distillation method is called pseudo-labeling \cite{kim-rush-2016-sequence}, where the teacher model generates pseudo summaries for all documents in the training set and the resulting document--\emph{pseudo}-summary pairs are used to train the student model.

\begin{figure}
    \centering
    \includegraphics[trim=85 20 10 10, clip, width=\linewidth]{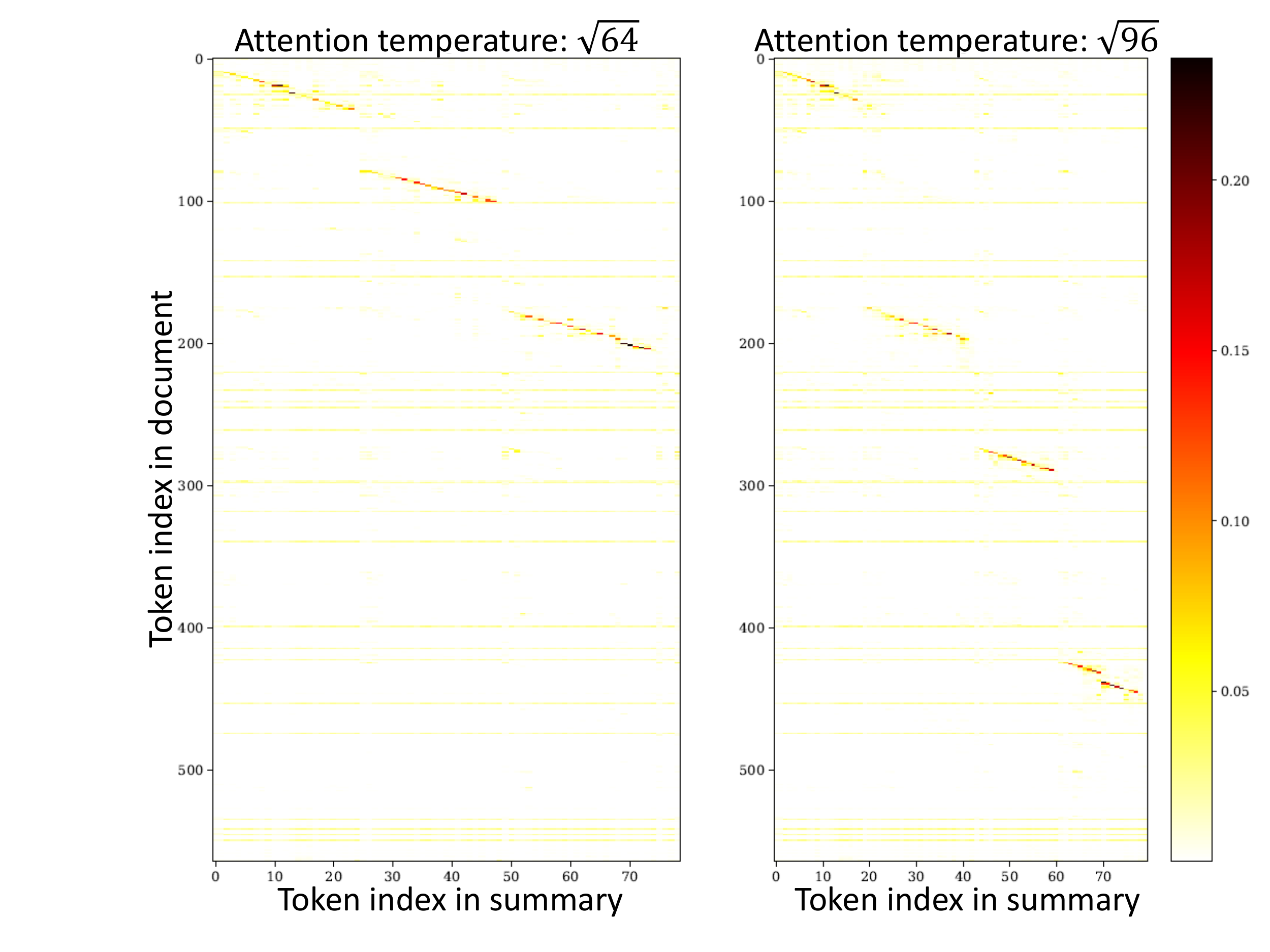}
    \caption{Visualization of teacher cross attention weights when generating pseudo labels with normal (left) and smoothed (right) attention weights. 
    This example is generated by the {\tt BART} teacher trained on CNNDM (see \S \ref{sec:results}). Its training and inference hyper-parameters are described in detail in \S \ref{MODEL_SETTINGS}.
    }
    \label{fig:pl_attn}
\end{figure}

In this paper, we argue that attention distributions of a Seq2Seq teacher model might be too sharp. As a result, pseudo labels generated from it are sub-optimal for student models. In the summarization task, we observe that 1) pseudo summaries generated from our teacher model copy more continuous text spans from original documents than reference summaries (56\% 4-grams in pseudo summaries and 15\% 4-grams in reference summaries are copied from their original documents on CNN/DailyMail dataset); 2) pseudo summaries tend to summarize the leading part of a document (measured on CNN/DailyMail, 74\% of sentences in pseudo summaries and 64\% of sentences in reference summaries are from the leading 40\% sentences in original documents). We obtain the two numbers above by matching each sentence in a summary with the sentence in its original document that can produce maximum ROUGE~\cite{lin-2004-rouge} score between them. We call the two biases above the \emph{copy bias} and the \emph{leading bias}. 
In order to have an intuitive feeling, we select a representative example~\footnote{See the detailed example in Appendix~\ref{EXAMPLE1}.} and visualize its cross attention weights~\footnote{We use cross attention because we can see how words in documents are selected during generation.} (see the left graph in Figure~\ref{fig:pl_attn}).
We observe that attention weights form three ``lines'', which indicates very time the decoder predicts the next word, 
its attention points to the next word in the input document. That may be the reason why multiple continuous spans of text are copied. Another phenomenon we observe is that all \emph{high-value} attention weights (in deeper color) concentrate on the first 200 words in the input document, which reflects the leading bias. In either case, the attention distribution is too sharp (i.e., attention weights of the next word position or the leading part is much larger than other positions), which means our teacher model is over-confident. 

Based on the observations above, we propose a simple method called {\sc Plate} (as shorthand for {\bf P}seudo-labeling with {\bf L}arger {\bf A}ttention {\bf TE}mperature) to smooth attention distributions of teacher models. Specifically, we re-scale attention weights in all attention modules with a higher temperature, which leads to  \emph{softer} attention distributions.
Figure \ref{fig:pl_attn} intuitively shows the effect of using higher attention temperatures. Compared with the left graph, the right graph with higher attention temperature has shorter lines (less copy bias) with high attention weights, and positions of high attention weights extend to the first 450 words (less leading bias).
Less \emph{copy bias} in pseudo summaries encourages student models to be more abstractive, while less \emph{leading bias} in pseudo summaries encourages student models to take advantage of longer context in documents.

Experiments on CNN/DailyMail, XSum, and New York Times datasets with student models of different sizes show {\sc Plate} consistently outperforms vanilla pseudo-labeling methods. Further empirical analysis shows that, with {\sc Plate}, both pseudo summaries generated by teacher models and summaries generated by student models are shorter and more abstractive, which matches the goal of abstractive summarization.

\section{Related Work}
Large pre-trained Seq2Seq Transformer models largely improve results of generation tasks including text summarization~\citep{song2019mass, lewis-etal-2020-bart, bao2020unilmv2, raffel2020exploring, zhang2020pegasus}. These models are pre-trained using unsupervised text-to-text objectives. For example, T5~\cite{raffel2020exploring} is pre-trained by predicting corrupted text spans. BART~\citep{lewis-etal-2020-bart} employs denoising auto-encoding objectives such as text infilling and sentence permutation during its pre-training. The pre-training objective of PEGASUS~\cite{zhang2020pegasus} is tailored for the summarization task, which predicts the most ``summary worthy'' sentences in a document. 
Our method aims to make these large models faster. 


In knowledge distillation, 
besides learning from gold labels in the training set, student models can learn from soft targets~\citep{ba2014deep, hinton2015distilling}, intermediate hidden states~\citep{romero2014fitnets}, attentions~\citep{DBLP:conf/iclr/ZagoruykoK17,wang2020minilm}, and target output derivatives~\citep{czarnecki2017sobolev} of teacher models. Recent work for distillation of pre-trained Transformers (e.g., DistilBERT~\citep{DBLP:journals/corr/abs-1910-01108}, TinyBERT~\citep{jiao-etal-2020-tinybert}, MobileBERT~\citep{sun-etal-2020-mobilebert}, BERT-of-Theseus~\citep{xu-etal-2020-bert}, MINILM~\citep{wang2020minilm}) focuses on natural language understanding tasks such as GLUE~\cite{wang-etal-2018-glue} or SQuAD~\cite{rajpurkar-etal-2016-squad} benchmarks. Most methods above are designed for classification models. 


In Seq2Seq learning tasks such as summarization, we can apply distillation methods above to each step of sequence model predictions. However, the sequence-level knowledge of teacher models is not well utilized. Therefore, \citet{kim-rush-2016-sequence} introduce a sequence-level knowledge distillation method (i.e., \emph{pseudo-labeling}), where a student model is trained with pseudo labels generated by the teacher model using beam search decoding. \citet{kim-rush-2016-sequence} and later work~\citep{kasai2020deep, gu2017non, denkowski-neubig-2017-stronger} show \emph{pseudo-labeling} achieves competitive performance for Seq2Seq tasks such as machine translation. \citet{shleifer2020pre} propose the \emph{shrink and fine-tune} (SFT) approach for pre-trained summarization distillation, which re-finetunes a teacher model with some layers removed, and they show SFT outperforms \emph{pseudo-labeling} and a modification of direct knowledge distillation \cite{jiao-etal-2020-tinybert} on one of their datasets, but not others. Our method, which builds on top of \emph{pseudo-labeling}, is conceptually simple and improves \emph{pseudo-labeling} across different summarization datasets.

There is an interesting line of work called self-distillation or self-training~\cite{furlanello2018born, xie2020self, deng2009imagenet, liu2020noisy, he2019revisiting}, where the size of the student model is identical to the size of the teacher model.
Our method can also be applied in self-distillation and can potentially be combined with the self-distillation methods above.

\section{Summarization Distillation}

\subsection{Transformer based abstractive summarization}
Abstractive summarization aims to rewrite a document into its shorter form (i.e., summary), which is a typical Seq2Seq learning problem.
We adopt the Seq2Seq Transformer \cite{vaswani2017attention} model. Given a document $X = (x_1, x_2, \dots, x_{|X|})$ and its gold summary $Y = (y_1, y_2, \dots, y_{|Y|})$, we estimate the following conditional probability:
\begin{equation}
	p(Y|X;\theta) = \prod_{t=1}^{|Y|} p(y_t|y_{<t}, X;\theta)
\end{equation}
where $\theta$ is the model parameter and $y_{<t}$ stands for all tokens before position $t$ (i.e., $(y_1, y_2, \dots, y_{t-1})$).

The Seq2Seq Transformer model can be trained by minimizing the negative log-likelihood of gold document-summary pairs:
\begin{equation}
    \mathcal{L}_{ \text{G} } (\theta) = - \frac{1}{|Y|} \log p(Y|X;\theta)
\end{equation}
where $|Y|$ is the number of tokens in summary $Y$.

\subsection{Distillation with pseudo labels}
\label{sec:distill}
Knowledge distillation refers to the task of transferring  knowledge of a large teacher model (or a group of large teacher models) into a small student model.
As to Seq2Seq learning tasks such as machine translation and summarization, pseudo-labeling based methods are usually used to imitate teacher predictions at the sequence level.
Specifically, suppose we have a document $X$, and $\hat{Y}=(\hat{y}_1, \hat{y}_2, \dots, \hat{y}_{|\hat{Y}|})$ is a \emph{pseudo} summary generated by a teacher model using beam search. The student can be trained by minimizing the negative log-likelihood of document-to-\emph{pseudo}-summary pairs.
\begin{equation}
    \mathcal{L}_{ \text{PL} } (\theta) = - \frac{1}{|\hat{Y}|} \sum_{t=1}^{|\hat{Y}|} \log p(\hat{y}_t|\hat{y}_{<t}, X;\theta)
\end{equation}
Strictly, all possible \emph{pseudo} summaries from $X$ should be taken into account. Unfortunately, the computational cost is prohibitive. We therefore use a single sample $\hat{Y}$ (which takes a large portion of probability mass from the teacher) instead as in \citet{kim-rush-2016-sequence}.



\subsection{Re-scaling attention temperatures}
\label{sec:rescale_attn_temp}
Both our teacher and student models are Seq2Seq Transformer  models. 
The core part of a Transformer model is the attention module:
\begin{equation}
    \mathrm{Attention}(Q, K, V) = \mathrm{softmax}(\frac{QK^T}{\tau})V
\end{equation}
where $Q$, $K$, $V$ are linear projections of hidden states of a layer and $\tau$ is the temperature of the attention module which is usually $\sqrt{d}$ ($d$ is the hidden dimension size of that attention head). 

Our distillation method {\sc Plate} works as follows. Assume we have a teacher model trained with $\tau = \sqrt{d}$. When the teacher generates pseudo labels with beam search, we use a higher attention temperature and set $\tau = \sqrt{\lambda \; d}$ where $\lambda > 1$ ($\lambda$ is the attention temperature coefficient). Note that we only change the teacher's attention temperature during inference time. When we train our student model with pseudo labels, we still use a normal temperature (i.e., $\tau = \sqrt{d}$). We find that adjusting the student's attention temperature does not work. Probably because the student can easily adapt to the scaled attention temperature during training. 

We find that $\lambda=1.5$ or $\lambda=2.0$ usually works well in practice. To encourage teacher models to generate pseudo labels with more diversity, we further propose to use a random $\lambda$ for each input document ($\lambda \sim U[a, b]$). Note that $U[a, b]$ is a uniform distribution and we typically set $a=1.0$ and $b=2.0$. 





\section{Experiments}

\subsection{Datasets}
We conduct our experiments on three popular document summarization datasets: CNN/DailyMail~\citep{hermann2015teaching}, XSum~\citep{narayan-etal-2018-dont}, and New York Times~\citep{sandhaus2008new}. 
All datasets are tokenized with the GPT-2 tokenizer \cite{radford2019language}, which is based on UTF-8 BPE~\cite{sennrich-etal-2016-neural}.

\paragraph{CNNDM} 
The CNN/DailyMail dataset (\mbox{CNNDM};~\citealp{hermann2015teaching}) contains online news articles from the CNN and DailyMail websites paired with their associated highlights as reference summaries. We follow the standard pre-processing steps described in \citet{see-etal-2017-get,liu-lapata-2019-text}.~\footnote{Scripts are available at~\url{https://github.com/abisee/cnn-dailymail}.}
The resulting numbers of document-summary pairs for training, validation, and test are 287,227, 13,368, and 11,490, respectively. 

\paragraph{XSum}
The XSum dataset is collected by harvesting online articles from the BBC with single sentence summaries, which is professionally written. The summaries are \emph{extremely} abstractive. 
We use the official splits of~\citet{narayan-etal-2018-dont}. There are 204,045 articles for training; 11,332 articles for validation; and 11,334 articles for test.
\paragraph{NYT}
The New York Times dataset (\mbox{NYT};~\citealp{sandhaus2008new}) is composed of articles published by New York Times, and the summaries are written by library scientists. 
After applying the pre-processing procedures described in ~\citet{durrett-etal-2016-learning, liu-lapata-2019-text}, we first obtain 110,540 articles with abstractive summaries. The test set is constructed by including the 9,076 articles published after January 1, 2007. The remaining 100,834 articles are further split into training and validation sets. After removing articles with summaries less than 50 words, we obtain the final dataset with 38,264 articles for training; 4,002 articles for validation; and 3,421 articles for test.

\subsection{Implementation details}

\begin{table}
\centering
\resizebox{\linewidth}{!}{
\begin{tabular}{@{}l|r|ccc@{}}
\toprule
\multirow{2}{*}{Model} & \multirow{2}{*}{\# Param.} & \multicolumn{3}{c}{Latency (Millisecond)} \\
                      &                           & CNNDM    & XSum    & NYT    \\ \midrule
\tt{BART}             & 406M                    & 1975     & 903     & 3272   \\
\tt{BART 12-6}              & 306M                     & 1279     & 438     & 1692   \\
\tt{BART 12-3}              & 255M                     & 924      & 289     & 1488   \\
\tt{Transformer}            & 70M                      & 1028     & 406     & 1462   \\ \bottomrule
\end{tabular}
}
\caption{Latency (in Milliseconds) on a V100 GPU and number of parameters (million) of our models.}
\label{tab:latency}
\end{table}

\paragraph{Teacher/Student model settings} \label{MODEL_SETTINGS}
We use {\tt BART} Large \cite{lewis-etal-2020-bart} as our teacher model, which has 12 layers in the encoder and decoder. The hidden size of each layer is 1024, and each layer contains 16 attention heads with a hidden size of 64. 
We have four kinds of student models. The first three student models are initialized from BART weights (therefore, their hidden sizes are the same as that of BART). All the three students have the 12 layers of BART encoder and differ in the number of decoder layers. They are denoted by {\tt BART 12-6}, {\tt BART 12-3}, and {\tt BART 12-12} with 6, 3, and 12 decoder layers, respectively.
For {\tt BART 12-6} (or {\tt BART 12-3}), the decoder is initialized from the first 6 (or 3) layers or the maximally spaced 6 (or 3) layers of {\tt BART} decoder.  
The fourth student is the Transformer base model~\citep{vaswani2017attention}, which has 6 layers in each of the encoder and decoder. Each layer has a hidden size of 512 and 8 attention heads. This student is randomly initialized and denoted by {\tt Transformer}. The latency statistics (Milliseconds) and numbers of parameters of above four models are in Table~\ref{tab:latency}.




\paragraph{Training and inference} 
Hyper-parameters for {\tt BART}, {\tt BART 12-6}, {\tt BART 12-3}, and {\tt BART 12-12} are similar. 
Specifically, all models are optimized using Adam \cite{kingma2014adam} with $\beta_1 = 0.9, \beta_2 = 0.999$. Learning rates are tuned on validation sets (choose from 1e-5, 3e-5, 5e-5, 7e-5). We truncate all documents and summaries to 1024 sub-word tokens. 
We use a batch size of around 80 documents (we limit the max number of tokens on each GPU to 2048) and train our models for 20,000/15,000/6,000 steps with 500 warmup steps for CNNDM, XSum, and NYT, respectively. We also employ a weight decay of 0.01. 
For {\tt Transformer}, the hyper-parameters of the Adam optimizer is a bit different, and we use $\beta_1 = 0.9, \beta_2 = 0.98$. Learning rates are picked from 1e-4, 3e-4, 5e-4, 7e-4 according to validation sets. The weight decay is set to 0.0001. The warmup step we use is 4000. We train {\tt Transformer} for 100 epochs and select the best model w.r.t. their ROUGE scores on validation sets. For all models above we apply a label smoothing of 0.1 to prevent overfitting \cite{DBLP:conf/iclr/PereyraTCKH17}.


During inference, as common wisdom, we apply beam search. The beam size, length penalty, and minimal length are 4, 2.0, and 55 on CNNDM; 6, 0.1, and 1 on XSum; and 4, 0.7, and 80 on NYT, respectively.
All our models are trained on 8 NVIDIA V100 GPUs. The training is fairly fast. Training on CNNDM with the teacher model (i.e., {\tt BART}) is most time-consuming. It takes about 45 minutes for one epoch, and we need 6 epochs in total. 

\subsection{Evaluations}
We evaluate the quality of different summarization systems using ROUGE. On CNNDM and XSum datasets, we report full-length F1 based ROUGE-1 (R1), ROUGE-2 (R2), and ROUGE-L (RL) scores. Following \citet{durrett-etal-2016-learning, liu-lapata-2019-text}, we report limited-length recall based ROUGE-1, ROUGE-2, and ROUGE-L, where generated summaries are truncated to the lengths of gold summaries. All ROUGE scores are computed using the {\tt ROUGE-1.5.5.pl} script~\footnote{with -c 95 -r 1000 -n 2 -a -m arguments.}. 

Summaries generated by abstractive models may be ungrammatical or unfaithful to the original document. Additionally, we also measure the quality of generated summaries by eliciting human judgements. We randomly sample 50 documents from the test set of CNNDM. 12 annotators are invited (they are either native English speakers or graduate students with IELTS test score over 6.5). In the evaluation, participants are presented with a document and a list of outputs by different models. First, they are asked to evaluate the summaries on three dimensions: \emph{fluency} (is the summary grammatically correct?), \emph{faithfulness} (is the summary faithful to the original document?), and \emph{coverage} (does the summary coverage important information of the document?). Then, they are asked to rank the summaries from best to worst as a way of determining the overall quality of summaries. Each document is ensured to be annotated by 3 different subjects.


\subsection{Results}
\label{sec:results}


\begin{table*}[h]
\centering
\resizebox{\linewidth}{!}{%
\begin{tabular}{l|c|lll|lll|lll} 
\toprule
\multicolumn{2}{c|}{\multirow{2}{*}{Model/Dataset}}               & \multicolumn{3}{c|}{CNNDM}                                                                                   & \multicolumn{3}{c|}{XSum}                                                                                   & \multicolumn{3}{c}{NYT}                                                                  \\
\multicolumn{2}{c|}{}                                             & \multicolumn{1}{c}{R1}      & \multicolumn{1}{c}{R2}      & \multicolumn{1}{c|}{RL}                          & \multicolumn{1}{c}{R1}      & \multicolumn{1}{c}{R2}                          & \multicolumn{1}{c|}{RL}     & \multicolumn{1}{c}{R1}      & \multicolumn{1}{c}{R2}      & \multicolumn{1}{c}{RL}       \\ 
\hline\hline
\multicolumn{11}{c}{Teacher}                                                                                                                                                                                                                                                                                                                                                              \\ 
\hline\hline
\multicolumn{2}{l|}{BERTSUM~\citep{liu-lapata-2019-text}}          & 42.13                       & 19.60                       & 39.18                                            & 38.81                       & 16.50                                           & 31.27                       & 49.02                       & 31.02                       & 45.55                        \\
\multicolumn{2}{l|}{T5-11B~\citep{raffel2020exploring}}            & 43.52                       & 21.55                       & 40.69                                            & \multicolumn{1}{c}{--}      & \multicolumn{1}{c}{--}                          & \multicolumn{1}{c|}{--}     & \multicolumn{1}{c}{--}      & \multicolumn{1}{c}{--}      & \multicolumn{1}{c}{--}       \\
\multicolumn{2}{l|}{PEGASUS~\citep{zhang2020pegasus}}              & 44.17                       & 21.47                       & 41.11                                            & 47.21                       & 24.56                                           & 39.25                       & \multicolumn{1}{c}{--}      & \multicolumn{1}{c}{--}      & \multicolumn{1}{c}{--}       \\
\multicolumn{2}{l|}{BART~\citep{lewis-etal-2020-bart}}             & 44.16                       & 21.28                       & 40.90                                            & 45.14                       & 22.27                                           & 37.25                       & \multicolumn{1}{c}{--}      & \multicolumn{1}{c}{--}      & \multicolumn{1}{c}{--}       \\
\multicolumn{2}{l|}{BART (ours)}                                  & 44.71                       & 21.52                       & 41.44                                            & 45.50                       & 22.26                                           & 36.98                       & 55.41                       & 36.59                       & 51.11                        \\ 
\hline\hline
\multicolumn{11}{c}{Student}                                                                                                                                                                                                                                                                                                                                                              \\ 
\hline\hline
\multicolumn{2}{l|}{BART-PL~\citep{shleifer2020pre}}               & \multicolumn{1}{c}{--}      & 19.93                       & \multicolumn{1}{c|}{--}                          & \multicolumn{1}{c}{--}      & 21.38                                           & \multicolumn{1}{c|}{--}     & \multicolumn{1}{c}{--}      & \multicolumn{1}{c}{--}      & \multicolumn{1}{c}{--}       \\
\multicolumn{2}{l|}{BART-KD~\citep{shleifer2020pre}}               & \multicolumn{1}{c}{--}      & 20.95                       & \multicolumn{1}{c|}{--}                          & \multicolumn{1}{c}{--}      & 21.63                                           & \multicolumn{1}{c|}{--}     & \multicolumn{1}{c}{--}      & \multicolumn{1}{c}{--}      & \multicolumn{1}{c}{--}       \\
\multicolumn{2}{l|}{BART-SFT~\citep{shleifer2020pre}}              & \multicolumn{1}{c}{--}      & 21.21                       & \multicolumn{1}{c|}{--}                          & \multicolumn{1}{c}{--}      & 21.08                                           & \multicolumn{1}{c|}{--}     & \multicolumn{1}{c}{--}      & \multicolumn{1}{c}{--}      & \multicolumn{1}{c}{--}       \\ 
\midrule
\multirow{5}{*}{\tt{BART 12-3}}       & Gold                & 44.28                       & 21.31                       & 41.18                                            & 44.33                            & 21.60                                                & 36.73                             & 54.75                       & 35.52                       & 50.56                        \\
                                      & Regular             & 43.65                       & 21.10                       & 40.40                                            & 44.40              & 21.63                                           & 36.44                       & 53.82                       & 35.12                       & 49.45                        \\
                                      & $\text{\sc Plate}_{\lambda=1.5}$             & 44.54\textbf{\textbf{$^*$}} & 21.70\textbf{\textbf{$^*$}} & 41.41\textbf{\textbf{$^*$}}                      & \textbf{44.40}                       & \textbf{21.92}                                  & \textbf{36.92$^*$}          & 54.47\textbf{\textbf{$^*$}} & 35.65                       & 50.39\textbf{\textbf{$^*$}}  \\
                                      & $\text{\sc Plate}_{\lambda=2.0}$             & \textbf{44.65$^*$}          & \textbf{21.78\textbf{$^*$}} & \multicolumn{1}{c|}{\textbf{41.71\textbf{$^*$}}} & 43.50                       & 21.45                                           & 36.47                       & \textbf{54.96\textbf{$^*$}} & \textbf{35.72}              & \textbf{51.05\textbf{$^*$}}  \\
                                      & $\text{\sc Plate}_{\text{rnd}}$ & 44.27\textbf{\textbf{$^*$}} & 21.50\textbf{\textbf{$^*$}} & 41.15\textbf{\textbf{$^*$}}                      & 44.21                       & 21.70                                           & 36.81\textbf{\textbf{$^*$}} & 54.60\textbf{\textbf{$^*$}} & 35.70                       & 50.53\textbf{\textbf{$^*$}}  \\ 
\midrule
\multirow{5}{*}{\tt{BART 12-6}}       & Gold                & 44.00                       & 21.08                       & 40.76                                            & 44.88                       & 21.75                                           & 36.72                       & 55.07                       & 35.91                       & 50.69                        \\
                                      & Regular             & 44.00                       & 21.08                       & 40.29                                            & 44.87                       & 21.65                                           & 36.47                       & 53.85                       & 35.08                       & 49.36                        \\
                                      & $\text{\sc Plate}_{\lambda=1.5}$             & 44.29$^*$                   & 21.57$^*$                   & 41.13\textbf{\textbf{$^*$}}                      & \textbf{45.13}              & 22.07\textbf{\textbf{$^*$}}                     & \textbf{37.13\textbf{$^*$}} & 54.41\textbf{\textbf{$^*$}} & 35.61\textbf{\textbf{$^*$}} & 50.29\textbf{\textbf{$^*$}}  \\
                                      & $\text{\sc Plate}_{\lambda=2.0}$             & \textbf{44.84\textbf{$^*$}} & \textbf{21.95\textbf{$^*$}} & \textbf{41.77\textbf{$^*$}}                      & 44.51                       & 21.79                                           & 36.92\textbf{\textbf{$^*$}} & \textbf{55.07\textbf{$^*$}} & \textbf{35.92\textbf{$^*$}} & \textbf{51.05\textbf{$^*$}}  \\
                                      & $\text{\sc Plate}_{\text{rnd}}$ & 44.38\textbf{\textbf{$^*$}} & 21.65\textbf{\textbf{$^*$}} & 41.27\textbf{\textbf{$^*$}}                      & 45.00                       & \textbf{22.09\textbf{$^*$}}                     & 37.09\textbf{\textbf{$^*$}} & 54.74\textbf{\textbf{$^*$}} & 35.88\textbf{\textbf{$^*$}} & 50.66\textbf{\textbf{$^*$}}  \\ 
\midrule
\multirow{4}{*}{\tt{BART 12-12}}      & Regular             & 43.58                       & 21.14                       & 40.33                                            & 44.55                       & 21.42                                           & 36.01                       & 54.36                       & 35.74                       & 49.97                        \\
                                      & $\text{\sc Plate}_{\lambda=1.5}$             & 44.72\textbf{\textbf{$^*$}} & 21.88\textbf{\textbf{$^*$}} & 41.55\textbf{\textbf{$^*$}}                      & \textbf{45.22\textbf{$^*$}} & \textbf{22.30\textbf{$^*$}}                     & \textbf{37.22\textbf{$^*$}} & 54.90                       & 36.17                       & 50.84\textbf{\textbf{$^*$}}  \\
                                      & $\text{\sc Plate}_{\lambda=2.0}$             & \textbf{45.08\textbf{$^*$}} & \textbf{21.98\textbf{$^*$}} & \textbf{42.07\textbf{$^*$}}                      & 44.76                       & 22.06\textbf{\textbf{$^*$}}                     & 37.09\textbf{\textbf{$^*$}} & \textbf{55.70\textbf{$^*$}} & \textbf{36.28}              & \textbf{51.70\textbf{$^*$}}  \\
                                      & $\text{\sc Plate}_{\text{rnd}}$ & 44.65\textbf{\textbf{$^*$}} & 21.80\textbf{\textbf{$^*$}} & 41.53\textbf{\textbf{$^*$}}                      & 44.60                       & 21.86\textbf{\textbf{$^*$}}                     & 36.69\textbf{\textbf{$^*$}} & 55.15\textbf{\textbf{$^*$}} & \textbf{36.28}              & 51.11\textbf{\textbf{$^*$}}  \\ 
\hline
\multirow{4}{*}{\tt{Transformer}} & Gold                & 40.29                       & 17.49                       & 36.71                                            & 29.04                       & \multicolumn{1}{c}{9.21}                        & 22.18                       & 49.44                       & 29.04                       & 45.07                        \\
                                      & Regular             & 41.00                       & 18.35                       & 37.65                                            & \textbf{30.19}              & \multicolumn{1}{c}{9.79}                        & 22.88                       & 49.97                       & 31.00                       & 45.88                        \\
                                      & $\text{\sc Plate}_{\lambda=1.5}$             & \textbf{41.19}              & 18.33                       & \textbf{38.01\textbf{$^*$}}                      & 29.40                       & \multicolumn{1}{c}{\textbf{10.11\textbf{$^*$}}} & \textbf{22.95\textbf{$^*$}} & 50.21                       & \textbf{31.14}              & 46.25                        \\
                                      & $\text{\sc Plate}_{\lambda=2.0}$             & 41.15                       & \textbf{18.41}              & 38.00\textbf{\textbf{$^*$}}                      & 28.56                       & 10.02\textbf{\textbf{$^*$}}                     & 22.83\textbf{\textbf{$^*$}} & \textbf{50.35}              & 30.75                       & \textbf{46.39}               \\
\bottomrule
\end{tabular}
}
\caption{Results of various models on CNNDM, XSum, and NYT datasets. ROUGE scores on CNNDM and XSum are F1 based and ROUGE scores on NYT are limited-length recall based. BART (ours) is our own implementation of BART fine-tuning. * indicates the model significantly outperforms the regular pseudo-labeling model (Regular).}
\label{tab:attn_temp_result}
\end{table*}
Our main results are shown in Table~\ref{tab:attn_temp_result}. The first block includes several recent abstractive summarization models based on large pre-trained Transformers. BERTSUM~\cite{liu-lapata-2019-text} employs BERT \cite{devlin-etal-2019-bert} as its encoder and uses randomly initialized decoder. 
T5~\cite{raffel2020exploring}, \mbox{PEGASUS}~\cite{zhang2020pegasus} and BART~\cite{lewis-etal-2020-bart} are three popular large Seq2Seq Transformer models with different pre-training objectives.
Our own fine-tuning version of BART (BART (ours)) is comparable or slightly better than the original reported BART results, and we use it as the teacher model on the three datasets.

The second block presents results of student models. ~\citet{shleifer2020pre} compare pseudo-labeling (BART-PL), knowledge distillation using both output and intermediate layers (BART-KD) as well as shrink and fine-tuning (BART-SFT) methods. They also use BART as teacher models. Note their settings of student models are {\tt BART 12-6} on CNNDM and {\tt BART 12-3} on XSum.

Results of our {\tt BART 12-3} and {\tt BART 12-6} student models are in the third and fourth block. We present results of students trained with gold labels (Gold) and regular pseudo labels (Regular) as well as pseudo labels with higher and random attention temperatures ($\text{\sc Plate}_{\lambda=1.5}^{\text{B12-3}}$, $\text{\sc Plate}_{\lambda=2.0}^{\text{B12-3}}$ and $\text{\sc Plate}_{\text{rnd}}^{\text{B12-3}}$). $\text{\sc Plate}_{\lambda=1.5}^{\text{B12-3}}$ means that the student uses attention temperature coefficient $\lambda=1.5$ with architecture setting \texttt{BART 12-3}. 
$\text{\sc Plate}_{\text{rnd}}^{\text{B12-3}}$ means that we use random attention temperature of $\lambda \sim  U[1.0, 2.0]$.
We observe that using pseudo-labeling methods with higher attention temperatures consistently improves over its counterpart with normal attention temperatures (Regular) across all three datasets, and the differences between them are almost always significant measured with the ROUGE script~\footnote{The script uses bootstrap re-sampling technology~\cite{davison1997bootstrap} to compute the 95\% confidence interval following~\citet{lin-2004-rouge}.} (see details in Table~\ref{tab:attn_temp_result}). 
Interestingly, our student models $\text{\sc Plate}_{\lambda=2.0}^{\text{B12-3}}$ and $\text{\sc Plate}_{\lambda=2.0}^{\text{B12-6}}$ outperform all models in comparison (including student models and even the teacher model) on CNNDM. Our best performing student model $\text{\sc Plate}_{\lambda=1.5}^{\text{B12-3}}$ outperforms BART-PL, BART-SFT, and BART-KD on XSum. Meanwhile, our method is conceptually simpler and can further be combined with their methods with additional training objectives. 

In Section \ref{sec:rescale_attn_temp}, we also propose a variant of our method, which employs random attention temperatures ($\text{\sc Plate}_{\text{rnd}}$ in Table~\ref{tab:attn_temp_result}). We can see that though random temperature based method is not as good as our best fixed-temperature method, it in general produces decent results. Therefore, we recommend using this method when the computing budget is limited. Note that we also tried more extreme $\lambda$ values as shown in Appendix~\ref{MORE_LAMBDA}, and we find the value of 1.5 or 2.0 works better than others.

In the fifth block, we additionally conduct self-distillation experiments, which is not the focus of this work. Our method improves the teacher model on CNNDM; ROUGE-2/L scores are improved on XSum; while on NYT, there are improvements on ROUGE-1/L.  

Results with the {\tt Transformer} student (the sixth block) follow a similar trend, although the improvements are smaller. It may because the modeling power of {\tt Transformer} without pre-training is not large enough to effectively model the differences in pseudo labels. It is also interesting to see that students distilled with pseudo-labeling do improve gold label based students using randomly initialized {\tt Transformer}, but not with pre-trained models (i.e., {\tt BART 12-6} and {\tt BART 12-3}), which may also be due to the strong modeling power of large pre-trained Transformers. 






\begin{table}
\centering
\resizebox{\linewidth}{!}{
\begin{tabular}{@{}lllll@{}}
\toprule
     & Ref                     & Regular           & $\text{\sc Plate}_{\lambda=1.5}^{\text{B12-6}}$           & $\text{\sc Plate}_{\lambda=2.0}^{\text{B12-6}}$  \\ \midrule
rank & \multicolumn{1}{c}{2.4} & \multicolumn{1}{c}{2.1} & \multicolumn{1}{c}{2.4} & \multicolumn{1}{c}{2.7$^*$} \\ \bottomrule
\end{tabular}
}
\caption{Human Evaluation on CNNDM dataset. * means significantly better than Regular.}
\label{tab:human_eval}
\end{table}

\paragraph{Human evaluation} \quad
We randomly sample 50 documents from the test set of CNNDM. We compare our best student model $\text{\sc Plate}_{\lambda=2.0}^{\text{B12-6}}$ against the regular pseudo-labeling model (Regular), another model $\text{\sc Plate}_{\lambda=1.5}^{\text{B12-6}}$ and human reference (Ref). We ask human judges to rank the outputs of these models from best to worst. We convert the ranks to rank ratings (rank $i$ to $5-i$) and further conduct student $t$-test on these ratings. As shown in Table \ref{tab:human_eval}, $\text{\sc Plate}_{\lambda=2.0}^{\text{B12-6}}$ obtains the best ranking score and the difference between $\text{\sc Plate}_{\lambda=2.0}^{\text{B12-6}}$ and the regular pseudo-labeling based method Regular is significant ($p < 0.05$), which indicates our proposed method $\textsc{Plate}$ indeed produces better summaries.

\begin{table}

\centering
\resizebox{\linewidth}{!}{
\begin{tabular}{@{}l|ccc@{}}
\toprule
Attention Setting                                & R1    & R2    & RL    \\ \midrule
$\lambda_{\text{enc}}=\lambda_{\text{cross}}=\lambda_{\text{dec}}=2.0$ & 45.65 & 22.59 & 42.60 \\
$\quad$ -- with $\; \lambda_{\text{enc}}=1.0$                  & 45.65 & 22.57 & 42.55 \\
$\quad$ -- with  $\; \lambda_{\text{cross}}=1.0$                & 44.45 & 21.52 & 41.22 \\
$\quad$ -- with $\; \lambda_{\text{dec}}=1.0$                  & 45.08 & 22.25 & 42.02 \\ \bottomrule
\end{tabular}
}
\caption{Effects of re-scaling attention temperatures for encoder self-attention, decoder self-attention, and decoder cross-attention on the validation set of CNNDM.}
\label{tab:ablation_study}
\quad

\end{table}

\paragraph{Ablation study}
In a Transformer, there are three types of attention modules (i.e., encoder self-attention, decoder self-attention and decoder cross-attention), and we can scale attention temperatures for all of them or some of them. 
Let $\lambda_{\text{enc}}$, $\lambda_{\text{cross}}$, and $\lambda_{\text{dec}}$ denote the attention temperature coefficient of the encoder self-attention module, the decoder cross-attention module, and the decoder self-attention module, respectively. 
As shown in Table~\ref{tab:ablation_study}, using large attention temperature coefficients (2.0) for all three types of attention modules leads to the best result. When setting the coefficient of the cross attention module to $\lambda_{\text{cross}} = 1.0$, the ROUGE scores drop most. 
Perhaps this is not surprising, since cross attentions are directly related to the selection of document contents for summarization. Besides, the attention temperature of the decoder self-attention is also crucial but not as important as the cross-attention (see the fourth row).

\begin{table}
\centering
\resizebox{\linewidth}{!}{
\begin{tabular}{@{}l|ccc@{}}
\toprule
Method   & R1    & R2    & RL    \\ \midrule
Sampling~\citep{edunov-etal-2018-understanding} & 43.70 & 20.83 & 40.56 \\
Nucleus Sampling & 43.86 & 20.95 & 40.68 \\
Output Layer $T=0.5$ & 43.80 & 21.20 & 40.59 \\
Regular  & 44.00 & 21.08 & 40.29 \\
$\text{\sc Plate}_{\lambda=2.0}$ (Ours)         & 44.84 & 21.95 & 41.77 \\ \bottomrule
\end{tabular}
}
\caption{Comparison with sampling and output layer temperature based distillation methods.}
\label{tab:SAMPLING}
\end{table}

\paragraph{Comparison with sampling and tuning output layer temperature} 
Sampling based methods can produce more diverse and richer outputs than its beam search based counterpart and has been proven useful in back translation~\citep{edunov-etal-2018-understanding}. We implement the sampling method in~\citet{edunov-etal-2018-understanding} and Nucleus Sampling~\citep{holtzman2019curious}, a more advanced sampling method, to generate pseudo labels for distillation. We use the {\tt BART 12-6} as the student model, and the distillation results on CNNDM are in Table~\ref{tab:SAMPLING}. As can be seen, both of the sampling based methods above perform worse than the regular beam search based pseudo-labeling method (Regular), let alone ours. Besides the attention temperatures, we can also tune the temperature $T$ in the decoder output softmax layer. With a proper $T$ (i.e., $T=0.5$) during pseudo label generation, the resulting student model slightly outperforms the baseline student model with regular pseudo labeling method on ROUGE-2/L (see Table~\ref{tab:SAMPLING}), but worse than $\text{\sc Plate}_{\lambda=2.0}$. More results with different $T$s are in Appendix~\ref{MORE_SOFTMAX_TEMP}.

\subsection{Analysis} \label{ANALYSIS}

\emph{Why does our distillation method work?} To answer this question, 
we first try to analyze the reasons from both the external characteristics of the summaries generated by the teacher model and the internal characteristics of the teacher's attention mechanism.
Then, we will give an in-depth explanation.

\begin{table*}[h]
\centering
\resizebox{\linewidth}{!}{
\begin{tabular}{@{}cccccccccccccc@{}}
\toprule
\multicolumn{2}{l}{}                                       & \multicolumn{4}{c|}{CNNDM}                          & \multicolumn{4}{c|}{XSum}                                   & \multicolumn{4}{c}{NYT}                 \\
\multicolumn{2}{c|}{$\lambda$ Setting}            & gold  & 1.0   & 1.5   & \multicolumn{1}{c|}{2.0}            & gold  & 1.0   & 1.5   & \multicolumn{1}{c|}{2.0}            & gold  & 1.0    & 1.5   & 2.0            \\ \midrule
\multicolumn{14}{c}{Average Length}                                                                                                                                                                                              \\ \midrule
Teacher                  & \multicolumn{1}{c|}{Avg. Len.} & 48.03 & 64.78 & 56.81 & \multicolumn{1}{c|}{\textbf{52.16}} & 21.10 & 20.33 & 17.28 & \multicolumn{1}{c|}{\textbf{15.66}} & 78.61 & 105.83 & 88.58 & \textbf{79.05} \\
Student                & \multicolumn{1}{c|}{Avg. Len.} & 67.51 & 82.31 & 73.10 & \multicolumn{1}{c|}{\textbf{65.92}} & 21.01 & 22.46 & 18.69 & \multicolumn{1}{c|}{\textbf{16.84}} & 92.61 & 109.78 & 98.16 & \textbf{88.52} \\ \midrule
\multicolumn{14}{c}{Novel $n$-grams Ratio(\%)}                                                                                                                                                                                     \\ \midrule
\multirow{4}{*}{Teacher} & \multicolumn{1}{c|}{1-gram}     & 25.24 & 7.89  & 9.15  & \multicolumn{1}{c|}{\textbf{12.56}} & 46.78 & 38.68 & 39.05 & \multicolumn{1}{c|}{\textbf{39.33}} & 12.96 & 4.04   & 4.34  & \textbf{6.25}  \\
                         & \multicolumn{1}{c|}{2-grams}    & 61.08 & 23.60 & 27.38 & \multicolumn{1}{c|}{\textbf{36.81}} & 87.83 & 80.50 & 81.91 & \multicolumn{1}{c|}{\textbf{82.70}} & 45.90 & 22.54  & 23.14 & \textbf{28.95} \\
                         & \multicolumn{1}{c|}{3-grams}    & 77.49 & 35.43 & 40.54 & \multicolumn{1}{c|}{\textbf{52.77}} & 97.17 & 93.09 & 94.27 & \multicolumn{1}{c|}{\textbf{94.91}} & 65.12 & 39.20  & 39.88 & \textbf{46.93} \\
                         & \multicolumn{1}{c|}{4-grams}    & 85.13 & 44.10 & 49.66 & \multicolumn{1}{c|}{\textbf{62.56}} & 99.08 & 96.78 & 97.64 & \multicolumn{1}{c|}{\textbf{98.07}} & 75.21 & 51.09  & 51.63 & \textbf{58.36} \\ \midrule
\multirow{4}{*}{Student} & \multicolumn{1}{c|}{1-gram}     & 23.55 & 4.58  & 5.07  & \multicolumn{1}{c|}{\textbf{6.56}}  & 46.80 & 37.33 & 38.01 & \multicolumn{1}{c|}{\textbf{38.07}} & 10.36 & 3.46   & 3.37  & \textbf{3.64}  \\
                         & \multicolumn{1}{c|}{2-grams}    & 58.52 & 15.16 & 16.64 & \multicolumn{1}{c|}{\textbf{21.40}} & 87.89 & 78.74 & 80.56 & \multicolumn{1}{c|}{\textbf{81.28}} & 41.16 & 21.21  & 20.50 & \textbf{21.93} \\
                         & \multicolumn{1}{c|}{3-grams}    & 75.50 & 24.36 & 26.58 & \multicolumn{1}{c|}{\textbf{33.67}} & 97.21 & 91.99 & 93.55 & \multicolumn{1}{c|}{\textbf{94.18}} & 60.65 & 37.60  & 36.67 & \textbf{38.71} \\
                         & \multicolumn{1}{c|}{4-grams}    & 83.49 & 31.70 & 34.36 & \multicolumn{1}{c|}{\textbf{42.74}} & 99.12 & 96.10 & 97.25 & \multicolumn{1}{c|}{\textbf{97.70}} & 71.48 & 49.56  & 48.47 & \textbf{50.56} \\ \bottomrule
\end{tabular}
}
\caption{Statistics on outputs of teachers and students with different attention temperature coefficient $\lambda$. The student models are all with the {\tt BART 12-6} setting.
Inference hyper-parameters on the same dataset are the same.
}
\label{tab:statistics}
\end{table*}

\paragraph{Length and novel $n$-grams} 
We first analyze the pseudo summaries generated by the teacher models. We calculate novel $n$-grams and lengths of generated summaries. Note that if an $n$-gram appears in the summary, but not in the original document, we call it a novel $n$-gram. Proportions of novel $n$-grams are used to measure the abstractiveness of summaries \cite{see-etal-2017-get,liu-lapata-2019-text}. As shown in Table~\ref{tab:statistics}, when using a larger $\lambda$, pseudo summaries are shorter~\footnote{{We also try changing the length penalty during teachers' inference to make pseudo summaries shorter, but we find this method does not help summarization distillation (see Appendix~\ref{EXP_LENGTH_PENALTY} for more details).}} and contain a larger portion of novel $n$-grams. It indicates that the teachers can produce more concise and abstractive summaries, which matches the goal of abstractive summarization.
\emph{Are these pseudo summaries of good quality?} The performance of the teacher with different attention temperatures on CNNDM test set is shown in Table \ref{tab:teacher-perf}. Their results are all decent and close to each other (at least for ROUGE-1 and ROUGE-L). Interestingly, compared with $\lambda=1.0$, the performance of the teacher with $\lambda=2.0$ is worse, but the resulting student is much better (see Table~\ref{tab:attn_temp_result}). Perhaps not surprisingly, the styles of summaries from students are similar with these from their teachers. Concise and abstractive teachers lead to concise and abstractive students (see Table~\ref{tab:statistics}). Conciseness and abstractiveness are good properties for summarization, which however may not be the case for other generation tasks such as machine translation. We apply {\sc Plate} to the WMT16~\citep{bojar-EtAl:2016:WMT1} English-German translation task and use Transformer-big as the teacher and Transformer-base as the student. With $\lambda=1.5$, we obtain a BLEU of 27.90, while the result of the regular pseudo-labeling is 27.79 (more details are in Appendix~\ref{MT}).

\begin{table}
\centering
\scalebox{0.9}{
\begin{tabular}{@{}c|ccc@{}}
\toprule
$\lambda$ & R1    & R2    & RL    \\ \midrule
1.0       & 44.71 & 21.52 & 41.44 \\
1.5       & 44.92 & 21.72 & 41.84 \\
2.0       & 44.38 & 21.02 & 41.50 \\ \bottomrule
\end{tabular}
}
\caption{ROUGE of teacher models with different attention temperature coefficient $\lambda$ on  test set of CNNDM.}
\label{tab:teacher-perf}
\end{table}



\paragraph{Attention} 

\begin{figure}
    \centering
    \includegraphics[trim=10 0 10 30,clip,scale=0.5]{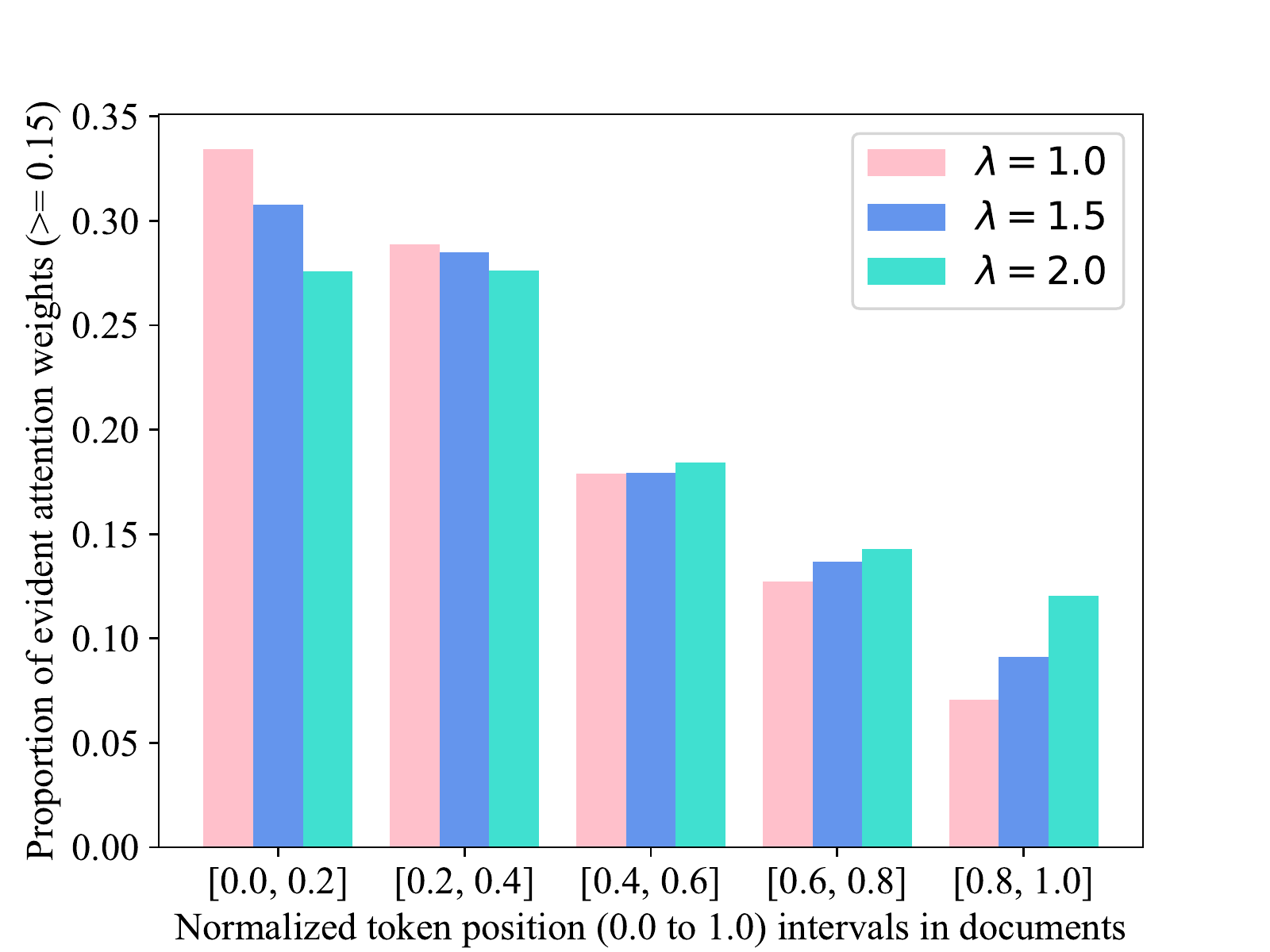}
    \caption{Distributions of \emph{evident} cross attention weights ($\ge 0.15$) when teachers generate pseudo labels with different attn. temperatures w.r.t. token positions.}
    \label{fig:teacher_attn_dist}
\end{figure}
We have shown earlier in Figure~\ref{fig:pl_attn} that with higher attention temperature, cross-attention modules of a teacher can attend to later parts in documents. We observe that students behave similarly, and we put more cross attention visualization of students in Appendix~\ref{ATTN_VIS}.
To obtain corpus-level statistics, we further calculate the \emph{evident} cross-attention weight distributions of the teacher when generating pseudo labels on the training set of CNNDM. Note that an attention weight is evident if it is greater than 0.15, and these evident attention weights account for around 15\% of all attention weights. Specifically, we normalize the token positions of each document to $(0.0, 1.0]$ and divide the normalized positions into five bins. The mean proportions of \emph{evident} attentions for all bins are shown in Figure~\ref{fig:teacher_attn_dist}. Compared to the teacher with normal attention temperature (pink bar), teachers with higher attention temperatures (blue and green bars) attend less on the heading parts of documents while more on the tail parts of documents. 
To sum up, teachers with higher attention temperatures can generate more concise and abstractive pseudo summaries, which makes the teacher provide more \emph{summary-like} pseudo labels to students. High-temperature teachers can alleviate the leading bias problems by providing pseudo labels with better coverage of source documents to students.

\paragraph{More explanation}
{According to the study of~\citet{xu-etal-2020-understanding-neural}, the prediction entropy correlates strongly with whether the model is copying or generating, as well as where in the sentence the token is (content selection). The decoder tends to copy when the model has a low prediction entropy and generate novel bigrams when the model has a high prediction entropy. They also find that high entropy of attention distribution strongly correlates with the model's high prediction entropy.}

{
Our method with a higher attention temperature makes attention distributions of the teacher model smoother and leads to a higher entropy of attention distributions, which results in a higher prediction entropy. Therefore, the model with higher attention temperature tends to copy less and generate more novel tokens. The conclusion from~\citet{xu-etal-2020-understanding-neural} is in accordance with our observation in Table~\ref{tab:statistics}.
}

\section{Conclusions}
In this work, we propose a simple but effective extension of pseudo-labeling method {\sc Plate} for summarization distillation. 
Experiments on three datasets demonstrate that our method can consistently outperform the vanilla pseudo-labeling method.
Further empirical analysis shows that by using our method, teacher models can generate more concise and abstractive summaries. As a result, summaries produced by student models also become more concise and abstractive. 
In the future, we would like to explore our method to other generation tasks as well as self-training with unlabeled data. We are also interested in combining our method with other distillation methods and extending our method for better teacher model training.

\bibliography{anthology}
\bibliographystyle{acl_natbib}

\include{acl_cr_appendix}

\end{document}

%% file: acl_cr_appendix.tex
\clearpage
\appendix

\section{Experiments of Applying \text{\sc Plate} to the Machine Translation Task} \label{MT}
We apply our method on the WMT16 En-De translation task. We use {\tt Transformer-Big} model as the teacher and {\tt Transformer-Base} as the student. Our results on newstest2014 are shown in Table~\ref{tab:MT_RESULT}. 
The student models with our method ($\lambda = 1.5$ and $\lambda = 2.0$) slightly outperform the student with regular pseudo-labeling method ($\lambda = 1.0$). Note that the improvement is not as significant as in summarization tasks. 

\begin{table}[]
\centering
\begin{tabular}{@{}lcc@{}}
\toprule
Model                    & $\lambda$ & BLEU  \\ \midrule
{\tt Transformer-Big} (teacher) & --        & 28.51 \\ \midrule
\multicolumn{3}{c}{Student}                  \\ \midrule
{\tt Transformer-Base}         & 1.0       & 27.79 \\
{\tt Transformer-Base}         & 1.5       & 27.90 \\
{\tt Transformer-Base}         & 2.0       & 27.85 \\ \bottomrule
\end{tabular}
\caption{Results of WMT En-De machine translation task on newstest2014. Student models are distilled from pseudo labels generated by the teacher with different attention temperatures ($\lambda$).}
\label{tab:MT_RESULT}
\end{table}

We speculate the reason may be that, unlike summarization, outputs of the machine translation task are relatively fixed. The strength of our method--conciseness and abstractiveness are good properties for summarization but seem not very beneficial to the translation task.

\section{Experiments of More $\lambda$ Values} \label{MORE_LAMBDA}
Besides the $\lambda$ values of 1.5 and 2.0, we also try more values in a broader range. Table~\ref{tab:MORE_LAMBDA} shows the distillation performance of {\tt BART 12-6} student models with more values of $\lambda$ we try on CNNDM dataset (we also include the values of 1.0, 1.5, and 2.0 in table for convenient comparison). As can be seen, both lower and larger $\lambda$ values are not helpful to the distillation. Though the suitable $\lambda$ values may vary across datasets, we recommend considering the $\lambda$ value 1.5 or 2.0 firstly in most cases.

\begin{table}[]
\centering
\begin{tabular}{@{}c|ccc@{}}
\toprule
$\lambda$ & R1    & R2    & RL    \\ \midrule
0.75      & 43.13 & 20.60 & 39.62 \\
1.0       & 44.00 & 21.08 & 40.29 \\
1.5       & 44.29 & 21.57 & 41.13 \\
2.0       & 44.84 & 21.95 & 41.77 \\
2.5       & 43.99 & 21.19 & 41.21 \\
3.0       & 42.32 & 19.28 & 39.67 \\ \bottomrule
\end{tabular}
\caption{ROUGE scores of {\tt BART 12-6} student models with more values of $\lambda$ on CNNDM dataset.}
\label{tab:MORE_LAMBDA}
\end{table}

\section{Experiments of Changing the Softmax Temperature in the Final Decoder Layer}
\label{MORE_SOFTMAX_TEMP}
It's a more direct idea to change the softmax temperature in the final decoder layer rather than attention temperatures, namely changing the $T$ in equation~\ref{SOFTMAX} to some other values rather than the default value 1.0. 
\begin{equation}
    q_i = \frac{\mathrm{exp}(z_i/T)}{\sum_j \mathrm{exp}(z_j / T)}
    \label{SOFTMAX}
\end{equation}

However, our experiments demonstrate that this method does not help summarization distillation much. We use {\tt BART} teacher models with different softmax temperatures in the final decoder layers to generate pseudo summaries and use the {\tt BART 12-6} as student models. The experiment results are shown in table~\ref{tab:SOFTMAX_TEMP}.
\begin{table}[]
\centering
\begin{tabular}{@{}c|ccc@{}}
\toprule
$T$       & R1    & R2    & RL    \\ \midrule
0.5     & 43.80 & 21.20 & 40.59 \\
1.0     & 44.00 & 21.08 & 40.29 \\
1.5     & 42.81 & 20.43 & 39.56 \\
2.0     & 42.76 & 20.34 & 39.53 \\ \midrule
Regular & 44.00 & 21.08 & 40.29 \\
$\text{\sc Plate}_{\lambda=2.0}$ (Ours)  & 44.84 & 21.95 & 41.77 \\ \bottomrule
\end{tabular}
\caption{Distillation experiment results of changing the softmax temperature in the final decoder layer.}
\label{tab:SOFTMAX_TEMP}
\end{table}

\section{Experiments of Shorter Pseudo Summaries with Smaller Length Penalty}
\label{EXP_LENGTH_PENALTY}

Our method can make pseudo summaries shorter and more abstractive, so one natural idea is that whether just changing the inference hyper-parameter length penalty to a smaller value, which can also make pseudo summaries shorter, can benefit abstractive summarization distillation. The experiment results are shown in Table~\ref{tab:LENGTH_PENALTY}, where the teacher is {\tt BART}, and the student is {\tt BART 12-6}.
As can be seen from the table, teachers with smaller length penalty (i.e., 1.0 or 0.5) cannot teach better students than the Regular pseudo-labeling or our method.

\begin{table}[t]
\centering
\resizebox{\linewidth}{!}{
\begin{tabular}{@{}c|c|ccc@{}}
\toprule
Length Penalty       & Avg. Len. & R1    & R2    & RL    \\ \midrule
1.0     & 64.39 & 42.96 & 20.67 & 39.76 \\
0.5     & 60.26 & 43.49 & 20.89 & 40.14 \\ \midrule
Regular & 64.78 & 44.00 & 21.08 & 40.29 \\
$\text{\sc Plate}_{\lambda=2.0}$ (Ours)  & 52.16 & 44.84 & 21.95 & 41.77 \\ \bottomrule
\end{tabular}
}
\caption{Distillation results of changing the teacher's inference hyper-parameter length penalty on CNNDM dataset. Avg. Len. represents the average length of the teacher generated pseudo summaries.}
\label{tab:LENGTH_PENALTY}
\end{table}

\section{The Example in Section~\ref{INTRO}} \label{EXAMPLE1}
We present the detailed content of the example in Section~\ref{INTRO} in table~\ref{tab:bias_example}.

\begin{table}[t]
    \centering
    \begin{tabular}{ p{0.95\linewidth} }
    \hline
    {\tt [Reference]}: Mentally ill inmates in Miami \textbf{are housed on the ``forgotten floor''} </s> Judge Steven Leifman says most are there as a result of \textbf{``avoidable felonies''} </s> While CNN tours facility, patient shouts: \textbf{``I am the son of the president}'' </s> Leifman says the system is unjust and he's fighting for change. \\
    \hline
    {\tt [PseudoLBL]}: Mentally ill inmates in Miami \textbf{are housed on the "forgotten floor"} of a \textbf{pretrial detention facility}. </s> Judge Steven \textbf{Leifman says about one-third of all people in Miami-Dade county jails are mentally ill}. </s> He says \textbf{they face drug charges or charges of assaulting an officer}, which are \textbf{``avoidable felonies''} </s> He says \textbf{the arrests often result from confrontations with police}, which \textbf{exacerbate their illness.} \\
    \hline
    {\tt [Smoothed ]}: Mentally ill inmates in Miami \textbf{are housed on the ``forgotten floor''} </s> Judge Steven Leifman says they are there because of \textbf{``avoidable felonies''} </s> He says many of them are in jail for drug or assault charges. </s> He says the system is unjust and he's trying \textbf{to change it.} \\
    \hline
    \end{tabular}
    \caption{Examples of reference summary ({\tt [Reference]}), pseudo summary from the teacher model ({\tt [PseudoLBL]}) and pseudo summary from the teacher with smoothed attention ({\tt [Smoothed ]}). Text spans in \textbf{bold} are copied spans (with more than four words) from the original document.}
    \label{tab:bias_example}
\end{table}

\section{Attention Visualization} \label{ATTN_VIS}
We present more examples of student models' outputs and cross attention visualization here. The student models are with the {\tt BART 12-6} setting and are trained on CNNDM and the following examples are from the validation set of CNNDM.

\begin{figure*}[h!]
    \centering
    \includegraphics[trim=130 50 140 50,clip,width=\linewidth]{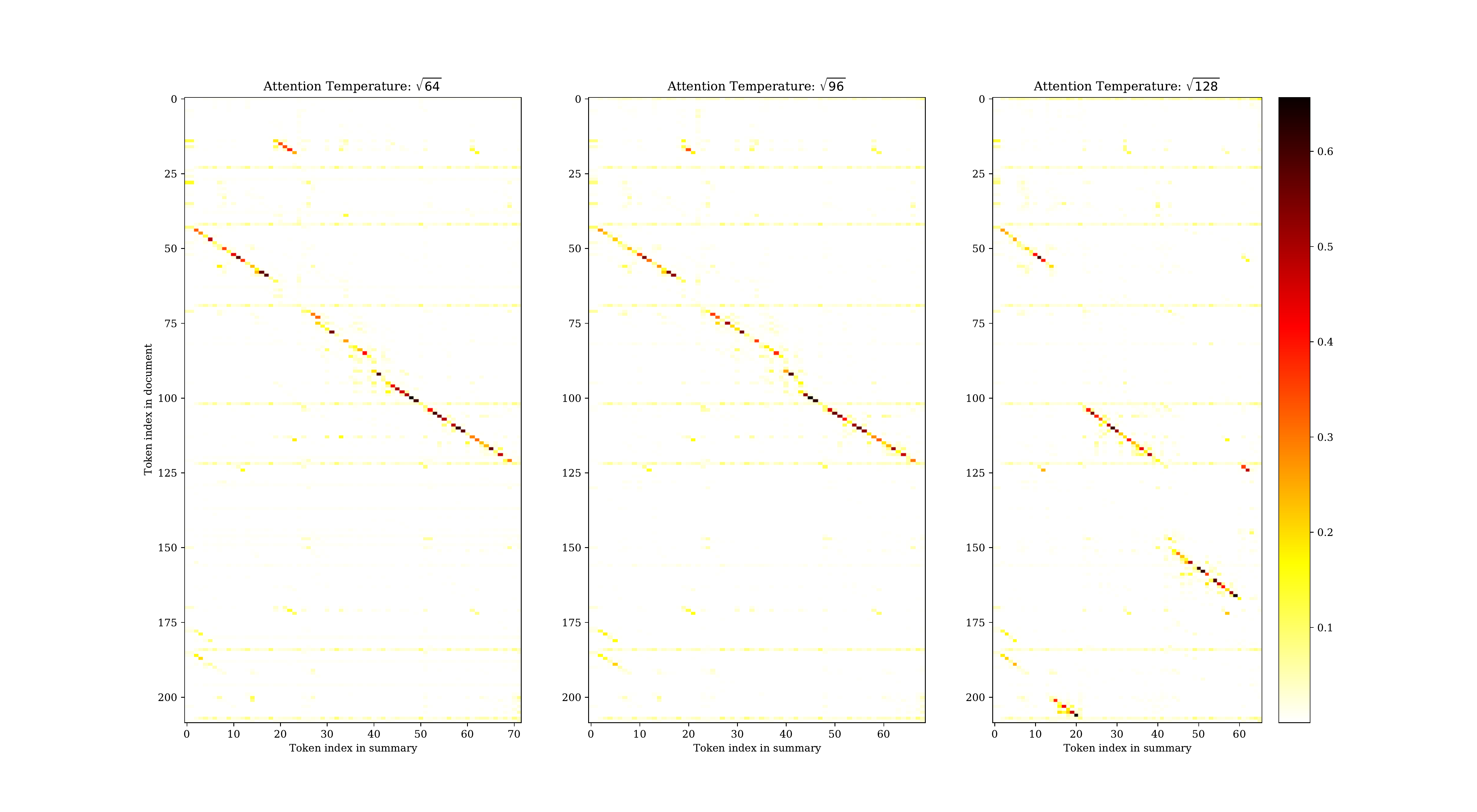}
    \caption{Example 1 of visualization of cross attention weight when the student generate summary with different attention temperatures.}
    \label{fig:student_attn_1}
\end{figure*}

\paragraph{Example 1} Table~\ref{tab:student_example_1} shows system outputs from different student models and Figure~\ref{fig:student_attn_1} illustrates the corresponding cross attention weights of these student models. Compared with the regular pseudo-labeling method ({\tt [Regular]}), the summary generated by our method $\text{\sc Plate}_{\lambda=1.5}^{\text{B12-6}}$ omits the modifier "Nirvana frontman" and "Nirvana bassist" of the person "Kurt Cobain" and "Krist Novoselic", respectively and the resulting summary is shorter and more abstractive. The summary generated by our method $\text{\sc Plate}_{\lambda=2.0}^{\text{B12-6}}$ contains the text "will premiere on HBO on May 4", which is at the end of the source document and included in the reference (i.e., summary worthy), but is ignored by {\tt [Regular]}. It indicates that our method can alleviate the leading bias problem. Figure~\ref{fig:student_attn_1} also shows that $\text{\sc Plate}_{\lambda=2.0}^{\text{B12-6}}$ can access the tail part of the document.
\begin{table}[h]
    \centering
    \begin{tabular}{ p{0.95\linewidth} }
    \hline
    {\tt [Reference]}: 
    The trailer surfaced on Wednesday. The documentary is winning raves. It \textbf{premieres on HBO on May 4}.
    \\
    \hline
    {\tt [Regular\quad]}:
    "Montage of Heck" is directed by Brett Morgen and offers an intimate portrayal of \textbf{Nirvana frontman Kurt Cobain}. The trailer features interviews with those closest to the performer, including family members, Courtney Love and \textbf{Nirvana bassist Krist Novoselic}. A companion book containing art and archival documents from Cobain is being produced to accompany the film.
    \\
    \hline
    {\tt [$\text{\sc Plate}_{\lambda=1.5}^{\text{B12-6}}$]}:
    "Montage of Heck" is directed by Brett Morgen and offers an intimate portrayal of \textbf{Kurt Cobain}. The trailer features bits of interviews with those closest to the performer, including family members, Courtney Love and \textbf{Krist Novoselic}. A companion book containing art and archival documents from Cobain is being produced to accompany the film.
    \\
    \hline
    {\tt [$\text{\sc Plate}_{\lambda=2.0}^{\text{B12-6}}$]}:
    "Montage of Heck" is directed by Brett Morgen and \textbf{will premiere on HBO on May 4}. A companion book containing art and archival documents from Cobain is being produced to accompany the documentary. The soundtrack will include "a mind-blowing 12-minute acoustic Cobain unheard track," Morgen says.
    \\
    \hline
    
    \end{tabular}
    \caption{Example 1 of reference summary ({\tt [Reference]}), summary generated from student with the regular pseudo-labeling method ({\tt [Regular]}), and summaries generated from students with {\sc Plate} ({\tt [$\text{\sc Plate}_{\lambda=1.5}^{\text{B12-6}}$]} and {\tt [$\text{\sc Plate}_{\lambda=2.0}^{\text{B12-6}}$]}.}
    \label{tab:student_example_1}
\end{table}

\begin{figure*}[]
    \centering
    \includegraphics[trim=130 50 140 50,clip,width=\linewidth]{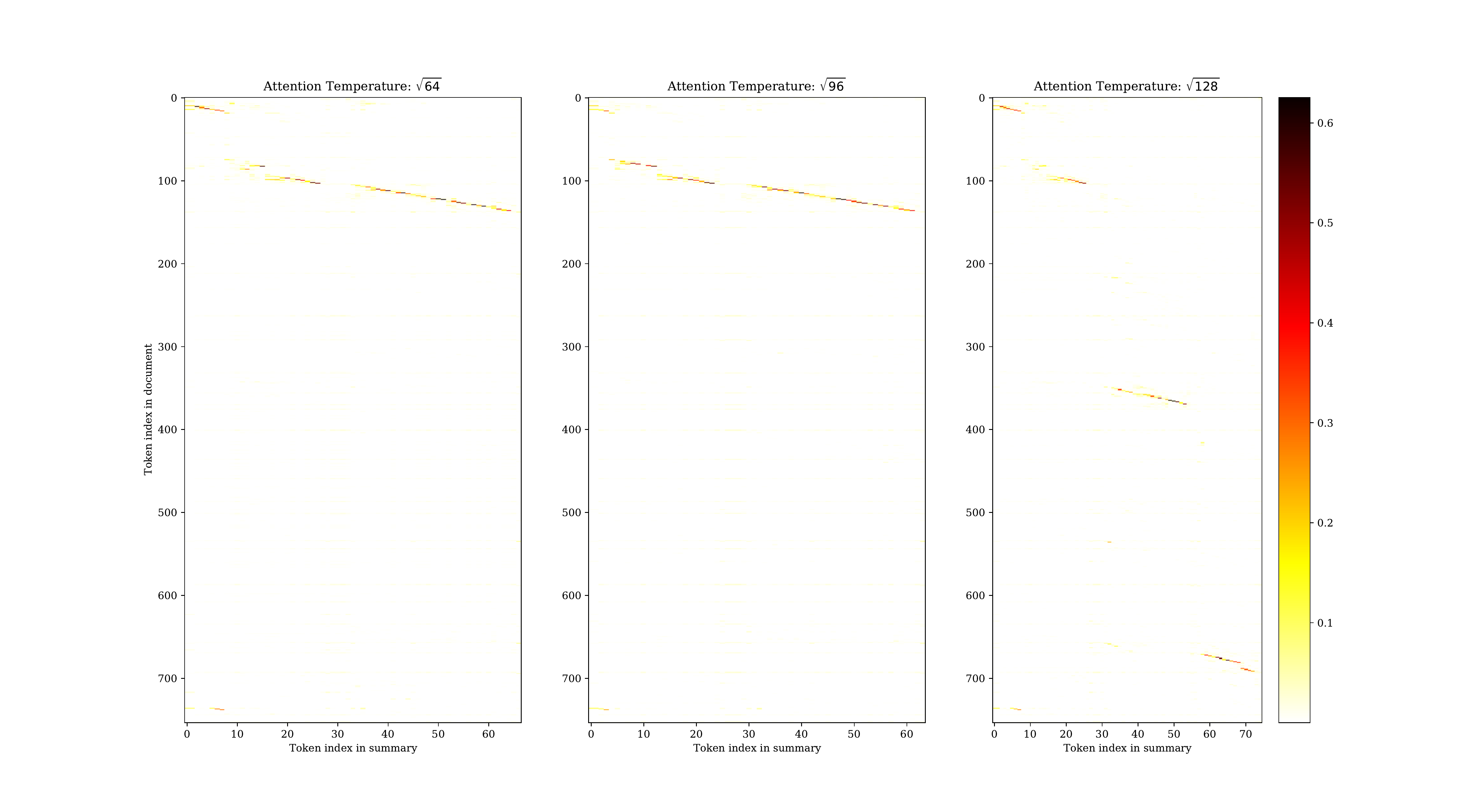}
    \caption{Example 2 of visualization of cross attention weight when the student generate summaries with different attention temperatures.}
    \label{fig:student_attn_3}
\end{figure*}

\paragraph{Example 2}
The second example is shown in Table~\ref{tab:student_example_3} (outputs) and Figure~\ref{fig:student_attn_3} (attention visualization).
In this example, the source document is relatively long (over 700 words). As shown in Figure~\ref{fig:student_attn_3}, the summary generated with the regular pseudo-labeling method {\tt Regular} mainly focuses on the heading part of the source document (around the first 150 words), but our method $\text{\sc Plate}_{\lambda=2.0}^{\text{B12-6}}$ takes into account the tokens in the front, middle and tail of the source document. In Table~\ref{tab:student_example_3}, the summary from $\text{\sc Plate}_{\lambda=2.0}^{\text{B12-6}}$ contains the key sentence "Peter Bergen: Pilots are not different from other people, but they can be careless, lazy, inattentive and reckless", which is similar to the reference sentence "Peter Garrison: Pilots don't exist on different moral plane than the rest of us". The sentence "the human mind is the blackest of boxes" in the reference, which appears at the tail of the source document, is also included in summaries of $\text{\sc Plate}_{\lambda=2.0}^{\text{B12-6}}$. This example again demonstrates that our method can alleviate the leading bias problem and can make the generated summary have better coverage of source documents.

\begin{table}[t]
    \centering
    \begin{tabular}{ p{0.95\linewidth} }
    \hline
    {\tt [Reference]}: 
    Experts suspect first officer Andreas Lubitz locked pilot out of the cockpit of plane. \textbf{Peter Garrison: Pilots don't exist on different moral plane than the rest of us, and the human mind is the blackest of boxes.}
    \\
    \hline
    {\tt [Regular\quad]}:
    Germanwings first officer Andreas Lubitz is one of a handful of airline pilots who have used their airplanes to combine suicide with mass murder. Frida Ghitis: Why is this thought at once so fascinating and so horrifying? It is because of the incompatibility between what we want to believe about flying and what we now see.
    \\
    \hline
    {\tt [$\text{\sc Plate}_{\lambda=1.5}^{\text{B12-6}}$]}:
    Andre Lubitz joins the short and infamous list of airline pilots who have used their airplanes to combine suicide with mass murder. Frida Ghitis: Why is this thought at once so fascinating and so horrifying? It is because of the incomp compatibility between what we want to believe about flying and what we now see.
    \\
    \hline
    {\tt [$\text{\sc Plate}_{\lambda=2.0}^{\text{B12-6}}$]}:
    Germanwings first officer Andreas Lubitz is one of a handful of pilots who have used their airplanes to combine suicide with mass murder. \textbf{Peter Bergen: Pilots are not different from other people, but they can be careless, lazy, inattentive and reckless. He says the human mind is the blackest of boxes; no one can peer inside it.}
    \\
    \hline
    \end{tabular}
    \caption{Example 2 of reference summary ({\tt [Reference]}), summary generated from student with the regular pseudo-labeling method ({\tt [Regular]}), and summaries generated from students with {\sc Plate} ({\tt [$\text{\sc Plate}_{\lambda=1.5}^{\text{B12-6}}$]} and {\tt [$\text{\sc Plate}_{\lambda=2.0}^{\text{B12-6}}$]}.}
    \label{tab:student_example_3}
\end{table}

